\documentclass{article} 
\usepackage[preprint]{colm2025_conference}

\usepackage{microtype}
\usepackage{hyperref}
\usepackage{url}
\usepackage{booktabs}
\usepackage{lineno}
\usepackage{graphicx}
\usepackage{subcaption}
\usepackage{minted}
\usepackage[most]{tcolorbox}
\usepackage{xspace}
\usepackage{xstring}
\usepackage[T1]{fontenc}
\usepackage{fvextra}
\usepackage{tikz}
\usetikzlibrary{tikzmark,calc,positioning,shadows}
\usepackage{booktabs}
\usepackage{multirow}
\usepackage{tabularx}   
\usepackage{siunitx}    
\usepackage{adjustbox}
\usepackage{array}

\definecolor{darkblue}{rgb}{0, 0, 0.5}
\hypersetup{colorlinks=true, citecolor=darkblue, linkcolor=darkblue, urlcolor=darkblue}

\usepackage{amsmath,amsfonts,bm}

\def\eqref#1{equation~\ref{#1}}

\def\1{\bm{1}}

\DeclareMathAlphabet{\mathsfit}{\encodingdefault}{\sfdefault}{m}{sl}
\SetMathAlphabet{\mathsfit}{bold}{\encodingdefault}{\sfdefault}{bx}{n}

\newcommand{\dllm}{\textbf{\texttt{dLLM}}\xspace}
\newcommand{\huggingface}{\raisebox{-1.5pt}{\includegraphics[height=1.05em]{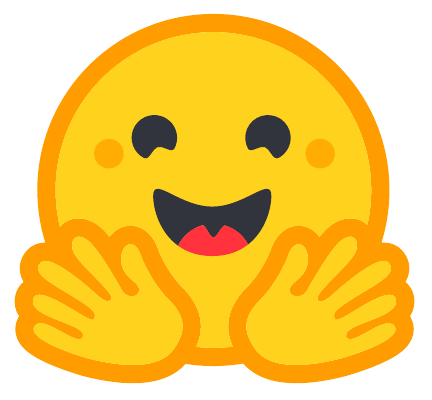}}\xspace}
\newcommand{\github}{\raisebox{-1.5pt}{\includegraphics[height=1.05em]{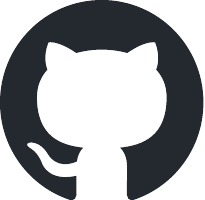}}\xspace}

\definecolor{vscbg}{HTML}{FFFFFF} 

\newtcolorbox{takeaway}[1][Takeaway]{%
  colback=black!4!white,
  colframe=black!60!white,
  fonttitle=\bfseries\small,
  colbacktitle=black!80!white,
  coltitle=white,
  enhanced,
  attach boxed title to top left={yshift=-2.2mm, xshift=4mm},
  boxrule=0.6pt,
  arc=1.5pt,
  left=6pt, right=6pt, top=6pt, bottom=4pt,
  title={#1},
}

\tcbset{
  vscodebox/.style={
    width=0.8\linewidth,      
    colback=vscbg,            
    boxrule=1pt,              
    arc=0pt,                  
    left=10pt, right=10pt,    
    top=3pt, bottom=0pt,      
    drop shadow={black!50!white}, 
    enhanced,                 
  }
}

\newenvironment{vscodebox}[1][]{
  \begin{tcolorbox}[vscodebox,#1]
}{
  \end{tcolorbox}
}

\newminted[pycode]{python}{
  fontsize=\small,    
  autogobble,         
  breaklines,         
  python3,            
  frame=none,         
  tabsize=4,          
  stripnl=false,      
  escapeinside=||,    
  texcomments=true,   
}

\definecolor{diffadd}{HTML}{22863A}      
\definecolor{diffdel}{HTML}{CF222E}      
\definecolor{diffaddbg}{HTML}{E6FFEC}    
\definecolor{diffdelbg}{HTML}{FFECEC}    

\newcommand{\DiffAdd}[1]{%
  {%
    \setlength{\fboxsep}{0pt}%
    \colorbox{diffaddbg}{%
      \makebox[\linewidth][l]{%
        \strut\textcolor{diffadd}{{\ttfamily #1}}%
      }%
    }%
  }%
}

\newcommand{\DiffDel}[1]{%
  {%
    \setlength{\fboxsep}{0pt}%
    \colorbox{diffdelbg}{%
      \makebox[\linewidth][l]{%
        \strut\textcolor{diffdel}{{\ttfamily #1}}%
      }%
    }%
  }%
}

\newcommand{\inlinecode}[1]{%
  \begingroup
  \setlength{\fboxsep}{1pt}%
  \colorbox{gray!15}{\smash[b]{\texttt{#1}}}%
  \endgroup
}

\title{\texttt{dLLM}: Simple Diffusion Language Modeling}

\author{Zhanhui Zhou\thanks{Equal contribution. Correspondence to \texttt{zhanhui@berkeley.edu}.} \\
UC Berkeley 
\And
Lingjie Chen$^*$ \\
UIUC
\And
Hanghang Tong \\
UIUC
\And
Dawn Song \\
UC Berkeley
}

\begin{document}

\ifcolmsubmission
\linenumbers
\fi

\maketitle

\begin{abstract}

Although diffusion language models (DLMs) are evolving quickly, many recent models converge on a set of shared components. These components, however, are distributed across ad-hoc research codebases or lack transparent implementations, making them difficult to reproduce or extend. As the field accelerates, there is a clear need for a unified framework that standardizes these common components while remaining flexible enough to support new methods and architectures.

To address this gap, we introduce \dllm, an open-source framework that unifies the core components of diffusion language modeling---\textbf{training, inference, and evaluation}---and makes them easy to customize for new designs. With \dllm, users can reproduce, finetune, deploy, and evaluate open-source large DLMs such as LLaDA and Dream through a standardized pipeline.
The framework also provides minimal, reproducible recipes for building small DLMs from scratch with accessible compute---including converting any BERT-style encoder or autoregressive LM into a DLM. We also release the checkpoints of these small DLMs to make DLMs more accessible and accelerate future research.
\newline

\hspace{5em} \github \dllm: \textbf{\url{https://github.com/ZHZisZZ/dllm}}

\hspace{5em} \huggingface \textbf{dllm-hub}: \textbf{\url{https://huggingface.co/dllm-hub}}
\end{abstract}

\section{Introduction}
\label{sec:introduction}

Diffusion language models (DLMs) have emerged as a promising alternative to standard autoregressive language modeling~\citep{austin2021structured,lou2023discrete,sahoo2024simple,shi2024simplified,arriolablock}, enabling iterative refinement~\citep{wang2025remasking,havasi2025edit}, flexible steering~\citep{li2022diffusion,schiff2025guidance} and efficient decoding~\citep{wu2025fastdllm,wu2025fastdllmv2,ma2025dkv,ben2025accelerated}. Alongside this rapid progress, a growing number of open-weight DLMs have appeared~\citep{nie2024scaling,nielarge,ye2025dream,rnd1_2025,bie2025llada2}, and many of them share similar design choices. However, these common components are frequently distributed across ad-hoc research codebases, or lack transparent implementations, making them difficult to reproduce, compare, or extend.


To address this critical gap, we introduce \dllm, an open-source framework that standardizes the end-to-end development pipeline for diffusion language modeling around three core components: \textbf{training}, \textbf{inference}, and \textbf{evaluation}. (1) For training, \dllm provides unified trainer modules that cover the most common objectives in DLMs, including Masked Diffusion~\citep{sahoo2024simple} and Block Diffusion~\citep{arriolablock}, while keeping diffusion modeling logic decoupled from model architectures so that new objectives and variants can be added with minimal refactoring.
In practice, this enables users to reproduce and finetune existing DLMs (e.g., LLaDA~\citep{nielarge} and Dream~\citep{ye2025dream}) and develop new models from scratch.
(2) For inference, \dllm introduces a lightweight abstraction that enables plug-and-play inference algorithms (including optimized efficient decoding algorithms~\citep{wu2025fastdllm}) without modifying existing model implementations. (3) For evaluation, \dllm provides a unified evaluation interface for reproducing official results across models.

Beyond unifying existing DLM development pipelines, \dllm provides minimal, reproducible recipes for building small DLMs with accessible compute. These recipes include transparent end-to-end pipelines for converting existing LMs (e.g., BERT-style encoders~\citep{devlin2019bert} and autoregressive language models~\citep{gong2024scaling}) into DLMs. We release checkpoints for these small models to support future research.

The key contributions of this work are:
\begin{itemize}
    \item We introduce \dllm, an open-source framework that unifies the core components of diffusion language modeling---training, inference, and evaluation---in a standardized, modular and extensible workflow, enabling transparent development and faster iteration across new designs.
    \item We release minimal, end-to-end recipes and checkpoints for training small DLMs from scratch (e.g., converting BERT-style encoders and autoregressive LMs into DLMs), providing accessible starting points and baselines for future research.
\end{itemize}

\section{Preliminaries}
\label{sec:preliminaries}

We denote a sequence of discrete tokens as $x = (x^1, \dots, x^L) \in \mathcal{V}^L$, where $\mathcal{V}$ is a finite vocabulary. We introduce a continuous time variable $t \in [0, 1]$ and a special mask token $m \notin \mathcal{V}$. The clean data is denoted $x_0$, and $x_t$ represents the corrupted sequence at time $t$.

\paragraph{Discrete Diffusion.}
Discrete diffusion models~\citep{austin2021structured,sahoo2024simple,lou2023discrete} generate data by reversing a forward process that progressively destroys information. The forward process $q(x_t | x_0)$ adds noise (e.g., random masking) over time $t: 0 \to 1$, transforming the data into an uninformative state $x_1$. The generative reverse process $p_\theta(x_{s} | x_t)$ (where $s < t$) learns to denoise $x_t$ to recover $x_0$. Unlike continuous diffusion, the state space remains discrete.

\paragraph{Masked Diffusion (MDLM).} 
Masked Diffusion (MDLM)~\citep{sahoo2024simple,shi2024simplified} simplifies the forward process as an absorbing-state masking process. In the forward process, each token $x_0^i$ is independently masked with probability $t$, assuming linear schedule:
\begin{equation}
    q(x_t^i | x_0^i) = (1 - t) \mathbb{I}(x_t^i = x_0^i) + t \mathbb{I}(x_t^i = m),
\end{equation}
where $\mathbb{I}$ is the indicator function. The model $p_\theta(x_0 | x_t)$ is trained to predict the unmasked tokens at indices $\mathcal{M}_t$ where $x_t^i = m$. The training objective minimizes the negative log-likelihood of the clean tokens given the masked input, with a time-dependent reweighting (e.g., 1/t assuming linear schedule) to balance contributions across noise levels:
\begin{equation}
    \mathcal{L}_{\text{MDLM}} = \mathbb{E}_{t \sim \mathcal{U}(0,1), x_0} \left[ \frac{1}{t} \sum_{i \in \mathcal{M}_t} - \log p_\theta(x_0^i | x_t) \right].
\end{equation}

\paragraph{Block Diffusion (BD3LM).}
Block Diffusion (BD3LM)~\citep{arriolablock} combines autoregression with diffusion. The sequence $x$ is partitioned into $K$ non-overlapping blocks $B_1, \dots, B_K$. We write $x^{B_k}$ to denote the tokens in block $B_k$, and $x_t^{B_k}$ for its corrupted version at time $t$. The model generates blocks autoregressively, but each block is generated via a diffusion process conditioned on the clean history of previous blocks $x_{<B_k}$. 
The joint probability factorizes as $p_\theta(x) = \prod_{k=1}^K p_\theta(x^{B_k} \mid x^{<B_k})$. For a specific block $B_k$, the objective is the diffusion loss averaged over time, strictly applied to the current block tokens while freezing the history. We define $\text{mask}(B_k, t)$ as the set of masked indices within block $B_k$ at time $t$, i.e., $\text{mask}(B_k, t) := \mathcal{M}_t \cap B_k$:
\begin{equation}
    \mathcal{L}_{\text{BD3LM}} = \sum_{k=1}^K \mathbb{E}_{t\sim \mathcal{U}(0,1), x_0} \left[ \frac{1}{t} \sum_{i \in \text{mask}(B_k, t)} - \log p_\theta(x_0^i \mid x_{t}^{B_k}, x^{<B_k}) \right].
\end{equation}
This factorization allows the model to leverage cached key-values from previous blocks while generating the current block in parallel.

\section{\dllm Overview}
\label{sec:dllm-overview}

In this section, we provide an overview of the three core components of \dllm: Trainer (Section~\ref{sec:trainer}), Sampler (Section~\ref{sec:sampler}) and Evaluation (Section~\ref{sec:evaluation}).

\subsection{Trainer}
\label{sec:trainer}

\begin{figure}[!h]
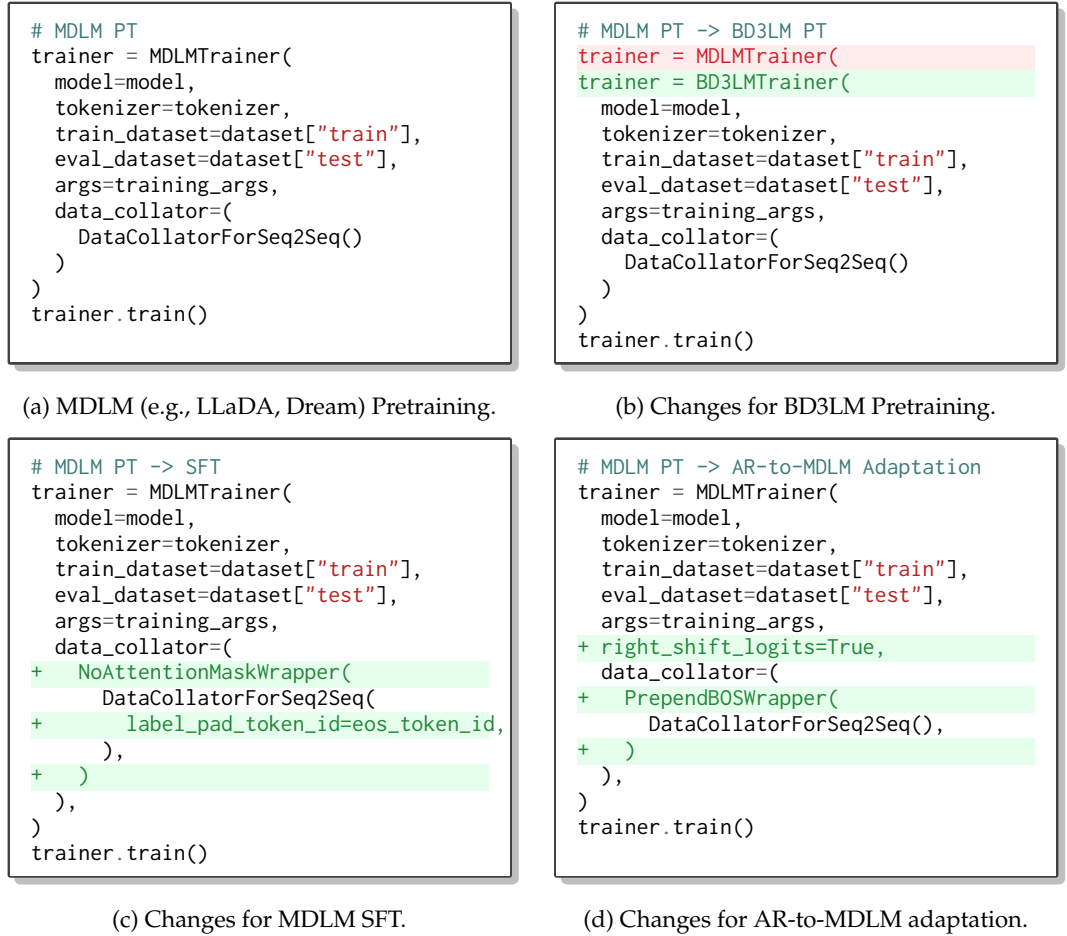

\centering

\begin{subfigure}{0.48\textwidth}
\centering
\begin{vscodebox}[width=\textwidth,left=5pt,right=5pt]
\begin{pycode}
# MDLM PT
trainer = MDLMTrainer(
  model=model,
  tokenizer=tokenizer,
  train_dataset=dataset["train"],
  eval_dataset=dataset["test"],
  args=training_args,
  data_collator=(
    DataCollatorForSeq2Seq()
  )
)
trainer.train()
|\phantom{ }|
\end{pycode}
\end{vscodebox}
\caption{MDLM (e.g., LLaDA, Dream) Pretraining.}
\label{fig:modular-trainer:mdlm-pt}
\end{subfigure}
\hfill
\begin{subfigure}{0.48\textwidth}
\centering
\begin{vscodebox}[width=\textwidth,left=5pt,right=5pt]
\begin{pycode}
# MDLM PT -> BD3LM PT
|\DiffDel{trainer = MDLMTrainer(}|
|\DiffAdd{trainer = BD3LMTrainer(}|
  model=model,
  tokenizer=tokenizer,
  train_dataset=dataset["train"],
  eval_dataset=dataset["test"],
  args=training_args,
  data_collator=(
    DataCollatorForSeq2Seq()
  )
)
trainer.train()
\end{pycode}
\end{vscodebox}
\caption{Changes for BD3LM Pretraining.}
\label{fig:modular-trainer:bd3lm-pt}
\end{subfigure}

\vspace{0.5em}

\begin{subfigure}{0.48\textwidth}
\centering
\begin{vscodebox}[width=\textwidth,left=5pt,right=5pt]
\begin{pycode}
# MDLM PT -> SFT
trainer = MDLMTrainer(
  model=model,
  tokenizer=tokenizer,
  train_dataset=dataset["train"],
  eval_dataset=dataset["test"],
  args=training_args,
  data_collator=(
|\DiffAdd{+~~~NoAttentionMaskWrapper(}|
      DataCollatorForSeq2Seq(
|\DiffAdd{+~~~~~~~label\_pad\_token\_id=eos\_token\_id,}|
      ),
|\DiffAdd{+~~~)}|
  ),
)
trainer.train()
\end{pycode}
\end{vscodebox}
\caption{Changes for MDLM SFT.}
\label{fig:modular-trainer:mdlm-sft}
\end{subfigure}
\hfill
\begin{subfigure}{0.48\textwidth}
\centering
\begin{vscodebox}[width=\textwidth,left=5pt,right=5pt]
\begin{pycode}
# MDLM PT -> AR-to-MDLM Adaptation
trainer = MDLMTrainer(
  model=model,
  tokenizer=tokenizer,
  train_dataset=dataset["train"],
  eval_dataset=dataset["test"],
  args=training_args,
|\DiffAdd{+~right\_shift\_logits=True,}|
  data_collator=(
|\DiffAdd{+~~~PrependBOSWrapper(}|
      DataCollatorForSeq2Seq(),
|\DiffAdd{+~~~)}|
  ),
)
trainer.train()
|\phantom{ }|
\end{pycode}
\end{vscodebox}
\caption{Changes for AR-to-MDLM adaptation.}
\label{fig:modular-trainer:mdlm-ar-to-mdlm}
\end{subfigure}

\caption{\textbf{A unified trainer interface supports a variety of purposes via modular trainers and configuration changes.}
Figure~\ref{fig:modular-trainer:mdlm-pt} shows the MDLM pretraining setup.
Figure~\ref{fig:modular-trainer:bd3lm-pt} shows the single-line trainer swap from \inlinecode{MDLMTrainer} to \inlinecode{BD3LMTrainer}.
Figure~\ref{fig:modular-trainer:mdlm-sft} shows the minimal changes to use \inlinecode{MDLMTrainer} for SFT: \inlinecode{NoAttentionMaskWrapper} keeps padding EOS visible, and \inlinecode{label\_pad\_token\_id=eos\_token\_id} trains the model to generate EOS from extra mask tokens in inputs.
Figure~\ref{fig:modular-trainer:mdlm-ar-to-mdlm} shows the minimal changes to adapt an autoregressive LM to MDLM: \inlinecode{right\_shift\_logits} reuses next-token prediction, and \inlinecode{PrependBOSWrapper} prepends BOS to provide the predictions for the first mask token.
}
\label{fig:modular-trainer}
\end{figure}

\paragraph{Unified training interface with Trainer (Figure~\ref{fig:modular-trainer}).}
Most open-weight DLMs to date are trained with Masked Diffusion (MDLM)~\citep{sahoo2024simple,nielarge,ye2025dream} or Block Diffusion (BD3LM)~\citep{arriolablock}. Accordingly, the current version of \dllm focuses on unified \inlinecode{MDLMTrainer} and \inlinecode{BD3LMTrainer} as core training modules (\href{https://github.com/ZHZisZZ/dllm/tree/main/dllm/core/trainers}{\inlinecode{dllm/core/trainers}}) that support both pretraining and finetuning most DLMs.
At the same time, the framework's modular design can be naturally extended to new diffusion objectives. For example, \dllm also includes a reference implementation of an EditFlow~\citep{havasi2025edit,nguyen2025oneflow} trainer for text diffusion with parallel insertion, substitution, and deletion operations.

\paragraph{Modular design enables easy customization (Figure~\ref{fig:modular-trainer}).}
Our training pipeline follows a modular design that allows core components to be reused and extended with minimal changes, improving both flexibility and readability. Figure~\ref{fig:modular-trainer} illustrates this modularity in practice: switching between MDLM/BD3LM pretraining, MDLM SFT, and AR-to-MDLM adaptation requires only localized changes (e.g., swapping the trainer, toggling a small set of arguments, or wrapping the data collator), without altering the overall pipeline.

\paragraph{Simple yet scalable training powered by \huggingface HF infrastructure.} 
Our training pipeline builds directly on the HuggingFace ecosystem. We use \href{https://github.com/huggingface/accelerate}{\inlinecode{accelerate}} to support diverse training configurations (e.g., FSDP ~\citep{zhao2023fsdp} and DeepSpeed~\citep{rajbhandari2020zero} for distributed training), and \href{https://github.com/huggingface/peft}{\inlinecode{peft}} for parameter-efficient finetuning. Our custom trainers (e.g., \inlinecode{MDLMTrainer}) are lightweight wrappers around the \href{https://github.com/huggingface/transformers}{\inlinecode{transformers}} Trainer~\citep{wolf2020transformers}. By using these components as building blocks, the framework stays easy to learn, which allows users to focus on DLM-specific logic and meanwhile remains scalable enough to support large-model pretraining and research experimentation.

\subsection{Sampler}
\label{sec:sampler}

\begin{figure}[!h]
\centering

\begin{subfigure}{0.78\textwidth}
\centering
\begin{vscodebox}[width=\textwidth]
\begin{pycode}
# Inference
|\DiffDel{sampler = MDLMSampler(model, tokenizer)}|
|\DiffAdd{sampler = MDLMFastdLLMSampler(model, tokenizer)}|

terminal_visualizer = TerminalVisualizer(tokenizer)

messages = [[{"role": "user", "content": "Write a python function."}]]

inputs = tokenizer.apply_chat_template(messages, add_generation_prompt=True, tokenize=True)

outputs = sampler.sample(inputs, return_dict=True)

terminal_visualizer.visualize(outputs.histories, rich=True)
\end{pycode}
\end{vscodebox}
\end{subfigure}

\caption{\textbf{Inference pipeline: sampler swap from vanilla to FastdLLM MDLM sampler.}}
\label{fig:inference-pipeline}
\end{figure}

\vspace{-1em}
\begin{figure}[!h]
\centering
\begin{tikzpicture}[
  imgnode/.style={inner sep=0, draw=black, thick,
    drop shadow={shadow xshift=3pt, shadow yshift=-3pt, opacity=0.5}}
  ]
  \node[anchor=south west, imgnode] (base) at (0,0)
    {\includegraphics[width=0.8\linewidth]{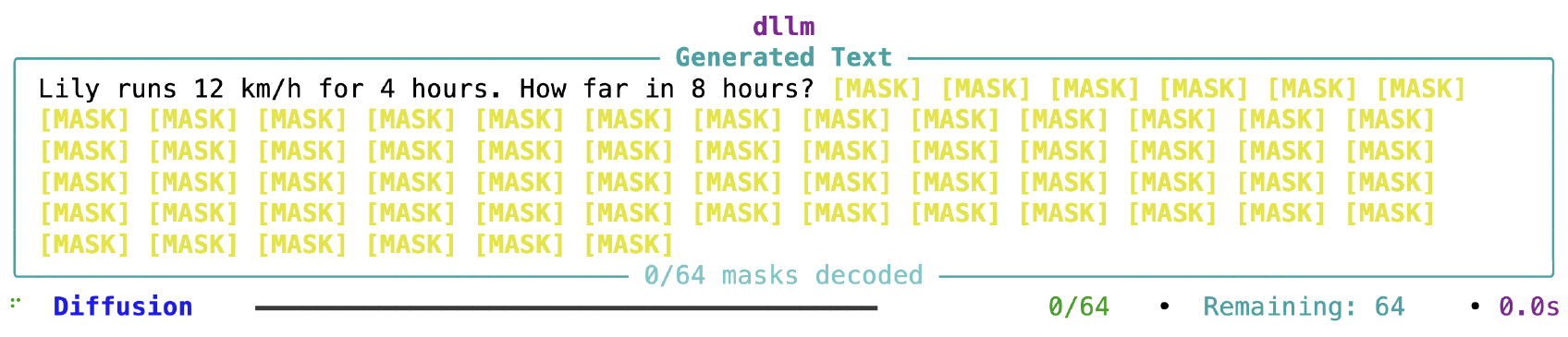}};
  \begin{scope}[x={(base.south east)}, y={(base.north west)}]
    \node[anchor=south, imgnode] at (0.5, -0.88)
      {\includegraphics[width=0.8\linewidth]{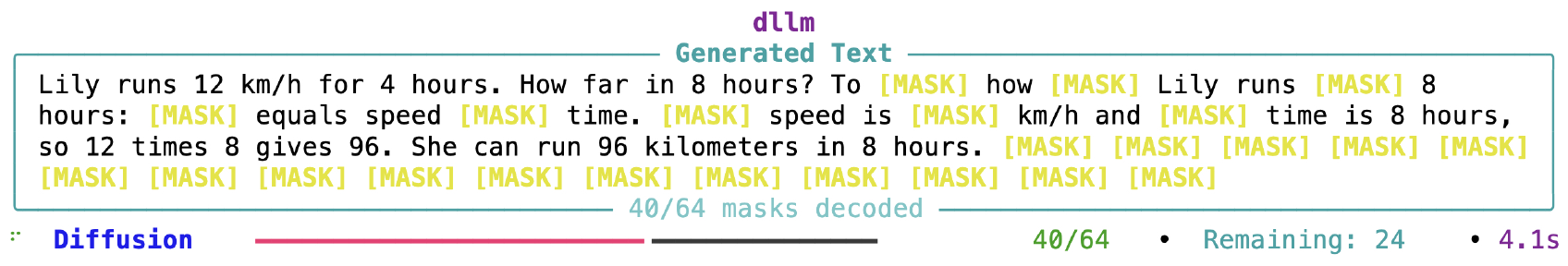}};
    \node[anchor=south, imgnode] at (0.5, -1.76)
      {\includegraphics[width=0.8\linewidth]{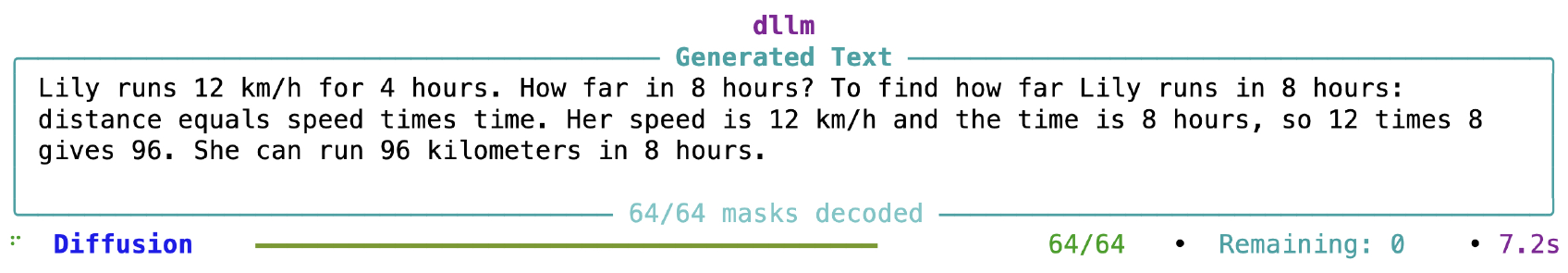}};
  \end{scope}
\end{tikzpicture}
\caption{\textbf{Terminal Visualizer showing transition from masked to decoded tokens.}}
\label{fig:terminal-visualization}
\end{figure}

\paragraph{Unified inference interface with Sampler (Figure~\ref{fig:inference-pipeline}).}
Different DLMs and inference algorithms expose inconsistent inference APIs, making it hard to reuse and compare inference algorithms across models.
To address this issue without modifying existing model implementations, we introduce a lightweight inference abstraction: \inlinecode{Sampler(model).sample()}.
This wrapper decouples models from inference algorithms, allowing different samplers to be swapped in a plug-and-play manner while keeping the underlying model unchanged.
Figure~\ref{fig:inference-pipeline} illustrates the unified inference pipeline enabled by this interface.

\paragraph{Terminal visualizer (Figure~\ref{fig:terminal-visualization}).}
Unlike autoregressive LMs, which decode tokens strictly left-to-right, DLMs decode tokens in any order.
As a result, the decoding order, beyond the final decoded output, is an important feature of DLMs and is valuable for analysis.
To support debugging and interpretability, we provide a terminal visualizer that reveals the token decoding order and the evolution of the sample over decoding steps (Figure~\ref{fig:terminal-visualization}).

\begin{figure}[htbp]
    \centering
    \begin{subfigure}[t]{\linewidth}
        \centering
        \includegraphics[width=\linewidth,trim=0 4em 0 0,clip]{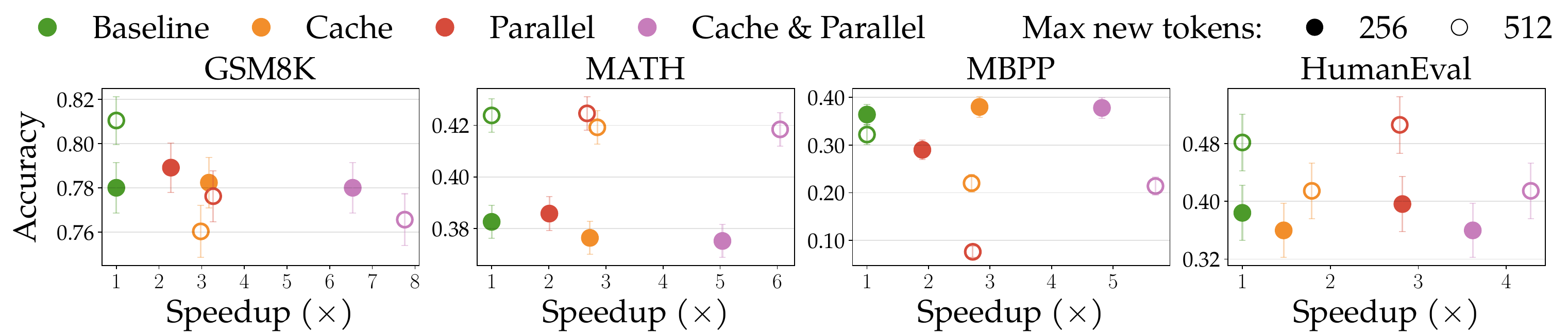}
        \caption{LLaDA-Instruct.}
        \label{fig:fastdllm-llada-eval}
    \vspace{0.5em}
    \end{subfigure}
    \begin{subfigure}[t]{\linewidth}
        \centering
        \includegraphics[width=\linewidth,trim=0 0 0 8em,clip]{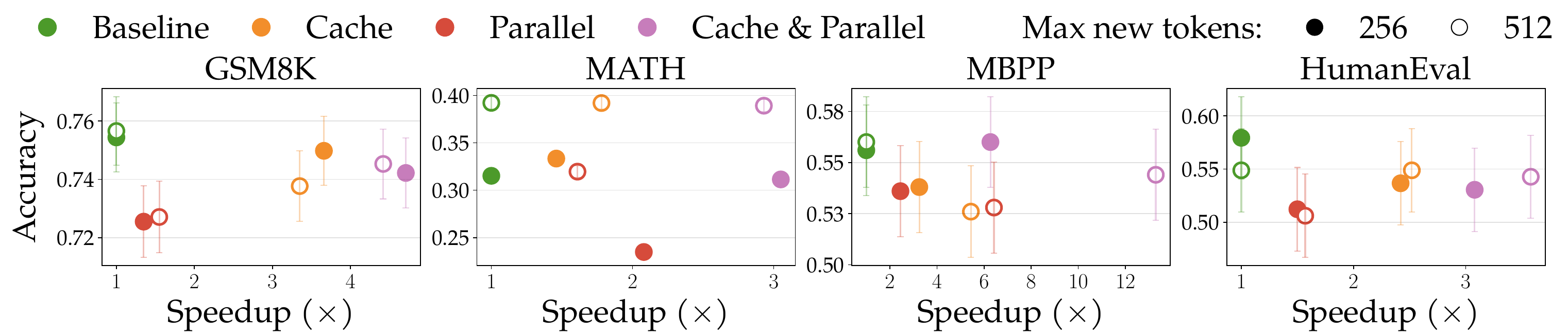}
        \caption{Dream-Base.}
        \label{fig:fastdllm-dream-eval}
    \end{subfigure}
    \vspace{-0.5em}
  \caption{\textbf{Fast-dLLM evaluation results with max new tokens @ $256$ and $512$.} Model selection follows the original Fast-dLLM evaluation for consistency and fair comparison. \textsc{Cache} uses block-wise approximate KV caching within each decoding block; \textsc{Parallel} uses confidence-based parallel token updates; \textsc{Cache \& Parallel} combines both. Note that max new tokens determines the number of pre-allocated padding tokens in the bidirectional context window, therefore affecting compute and measured performance.}

    \label{fig:fastdllm-eval}
\end{figure}

\paragraph{Efficient DLM inference (Figures~\ref{fig:inference-pipeline} \&~\ref{fig:fastdllm-eval}).}
DLM inference speed is a practical bottleneck~\citep{wu2025fastdllm,wu2025fastdllmv2,ma2025dkv,ben2025accelerated}. Building on the unified inference interface, \dllm includes an implementation of Fast-dLLM~\citep{wu2025fastdllm} for accelerated MDLM decoding: \inlinecode{MDLMFastdLLMSampler}, which can be used as a drop-in replacement for the standard \inlinecode{MDLMSampler} (Figure~\ref{fig:inference-pipeline}).
We report benchmarking results consistent with official Fast-dLLM implementations, demonstrating substantial inference speedups (see Figure~\ref{fig:fastdllm-eval} for visualization and Tables~\ref{tab:llada-fastdllm-eval} and~\ref{tab:dream-fastdllm-eval} in Appendix~\ref{app:sec:evaluation-reproduction} for detailed results).

\subsection{Evaluation}
\label{sec:evaluation}

\begin{figure}[t]
    \centering
    
    \begin{subfigure}[t]{0.48\linewidth}
        \centering
        \includegraphics[width=\linewidth]{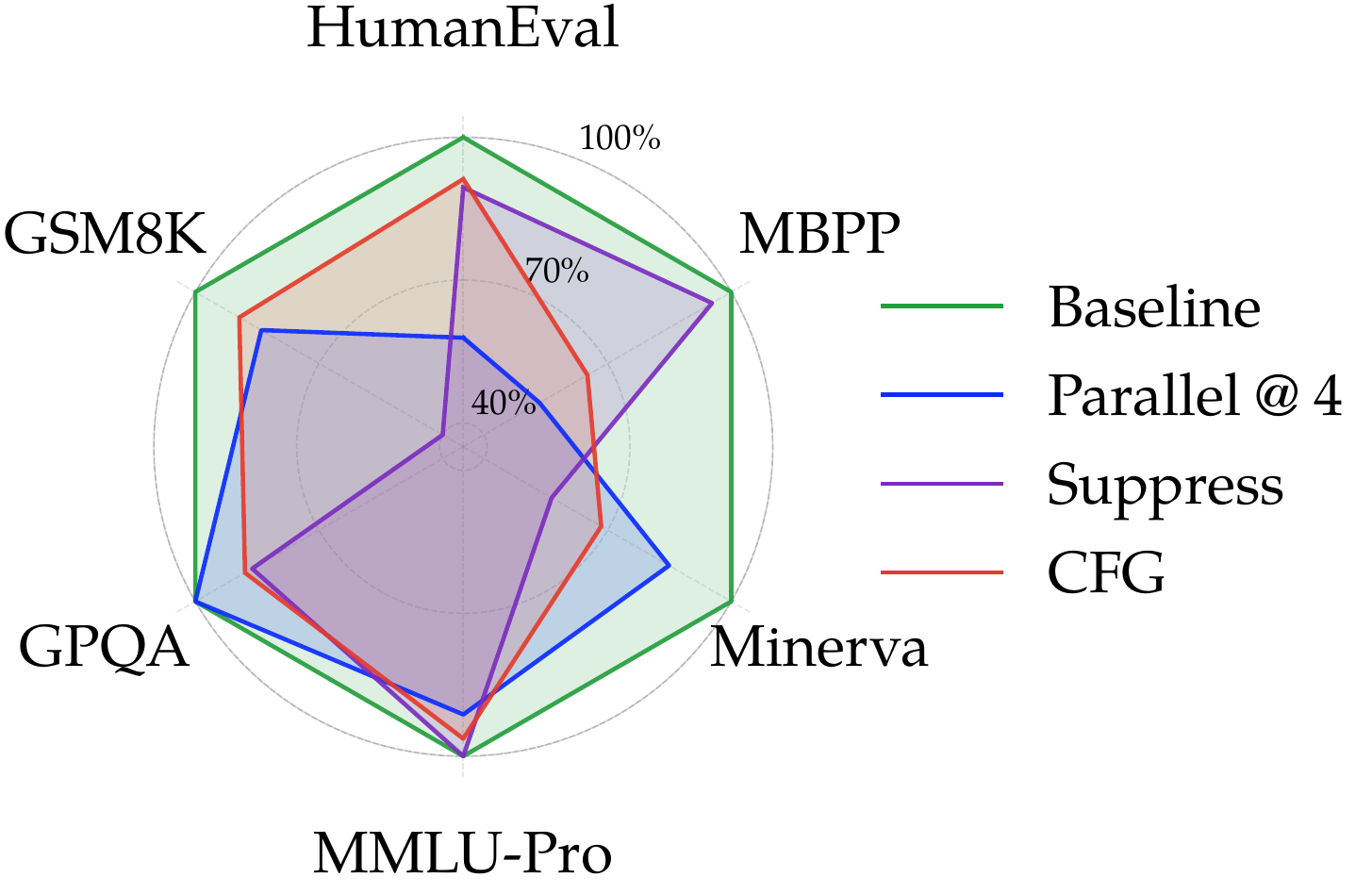}
        \caption{LLaDA-Instruct}
        \label{fig:temp-sensitivity}
    \end{subfigure}
    \hfill
    \begin{subfigure}[t]{0.48\linewidth}
        \centering
        \includegraphics[width=\linewidth]{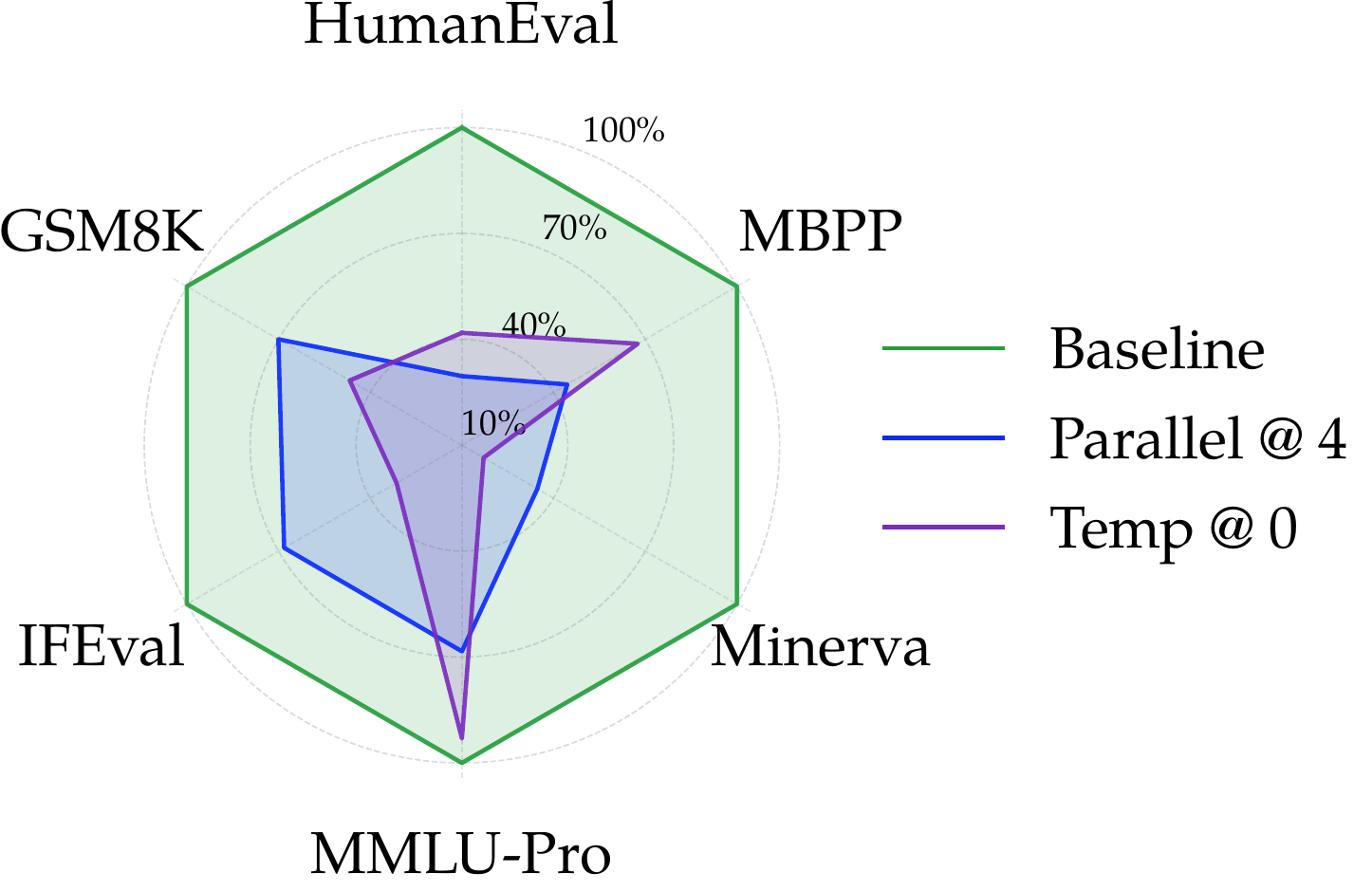}
        \caption{Dream-Instruct}
        \label{fig:topp-sensitivity}
    \end{subfigure}
    \caption{\textbf{Sensitivity to decoding hyperparameters.} 
    We vary individual sampling hyperparameters at inference time and observe that performance can degrade sharply from the optimal configuration. 
    \textsc{Baseline} denotes the best-performing setting; 
    \textsc{Suppress} does not suppress \texttt{<eos>} from the beginning of generation; 
    \textsc{CFG} sets \texttt{cfg=0.5}; 
    \textsc{Parallel @ $4$} generates four tokens per step; 
    and \textsc{Temp @ $0$} sets \texttt{temperature=0.0}.}
    \label{fig:hyperparam-sensitivity}
\end{figure}

Open-weight DLMs~\citep{nielarge,ye2025dream} rely on different evaluation tools, making unified evaluation difficult. This is further complicated by the fact that DLMs are especially sensitive to inference hyperparameters, as prior work often relies on task-specific hyperparameter tuning and postprocessing to achieve the best performance. For example, even a single change in an inference parameter can significantly alter performance (Figure~\ref{fig:hyperparam-sensitivity}).



A unified evaluation pipeline must therefore be flexible enough to support customization while faithfully reproducing the evaluation configurations used in prior work. To achieve this, we extend the \href{https://github.com/EleutherAI/lm-evaluation-harness}{\inlinecode{lm-evaluation-harness}}~\citep{eval-harness} framework and carefully match the preprocessing, decoding settings, and post-processing used for each model--task pair with its corresponding official pipeline. These details vary across models and tasks and require manual verification, but this enables our framework to reproduce the reported, model-specific scores while supporting consistent comparisons across models. Tables~\ref{tab:llada-eval} and~\ref{tab:dream-eval} (Appendix~\ref{app:sec:evaluation-reproduction}) compare our reproduced results against the originally reported results, showing that our evaluation framework closely matches the official results.

\begin{takeaway}
\textbf{DLM evaluation is highly sensitive to inference hyperparameters} (e.g., max new tokens in Figure~\ref{fig:fastdllm-eval} and other hyperparameters in Figure~\ref{fig:hyperparam-sensitivity}), a fact that is rarely explicitly stated in prior work, but is evident from our reproduction experiments.
\end{takeaway}

\section{Open DLMs with Open Recipes}
\label{sec:recipes}

Building on \dllm, we provide a set of fully reproducible recipes for training DLMs.
These recipes cover (1) finetuning open-weight DLMs to reason (Section~\ref{sec:recipes-finetuning-open-weight-large-dlms}), and (2) training small DLMs from scratch with minimal compute (e.g., Section~\ref{sec:recipes-bert-chat}, Section~\ref{sec:recipes-training-small-dlms}).
We make all of these model checkpoints available in \huggingface \href{https://huggingface.co/dllm-hub}{dllm-hub} along with their evaluation results.

\subsection{Finetuning Open-Weight Large DLMs}
\label{sec:recipes-finetuning-open-weight-large-dlms}

Training autoregressive LMs to reason before providing final answer has proven effective in solving complex tasks. Recent efforts such as d1~\citep{zhao2025d1} have begun exploring similar reasoning capabilities in DLMs. Using the unified trainer in \dllm, finetuning large DLMs is straightforward. We demonstrate that MDLM-style SFT can elicit reasoning capabilities in existing open-weight DLMs and improve their downstream performance.

\paragraph{Training details.}
We finetune both the Base and Instruct variants of LLaDA~\citep{nielarge} and Dream~\citep{ye2025dream} using MDLM SFT with LoRA on the s1K dataset~\citep{muennighoff2025s1}. Loss is computed only on response tokens. We use maximum sequence length of $4096$, $20$ epochs, learning rate $10^{-5}$, global batch size $32$ with gradient accumulation steps of $4$. We apply LoRA adaptation with $r=128$, $\alpha=256$, and weight decay $0.1$. We adopt a cosine learning-rate schedule with $10\%$ warmup. Training is conducted on $8 \times$A100 GPUs using DeepSpeed ZeRO-2. See Figure~\ref{fig:recipes-finetuning-open-weight-large-dlms} for training curves.

\paragraph{Evaluation results.}
We evaluate models SFTed on reasoning data by prepending a \texttt{<reasoning>} token at inference to force reasoning (Table~\ref{tab:lora-finetuning-results}) using evaluation pipelines from ~\cite{zhao2025d1}. For Instruct models, reasoning SFT yields consistent gains across math, planning, and coding benchmarks. Base models show improvements on in-distribution math tasks (e.g., GSM8K~\citep{cobbe2021gsm8k}, MATH500~\citep{hendrycks2021math}) but regress on out-of-distribution benchmarks. Overall, the results indicate that SFT is an effective starting point for reasoning in DLMs; all of these are achieved with the unified trainer interface (Figure~\ref{fig:modular-trainer}) with little changes.

\begin{table*}[!t]
\centering
\small
\setlength{\tabcolsep}{6pt}
\begin{tabular}{lcccccc}
\toprule
Model
& GSM8K
& MATH500
& Countdown
& Sudoku
& HumanEval
& MBPP\\
\midrule

LLaDA-Instruct
& $79.91$
& $34.80$
& $19.92$
& $11.62$
& $35.37$
& $42.02$ \\
+ SFT (MDLM)
& $80.59$
& $35.40$
& $26.95$
& $14.36$
& $36.59$
& $43.97$ \\

\midrule

LLaDA-Base
& $64.67$
& $10.20$
& $10.16$
& $0.34$
& $25.00$
& $40.08$ \\
+ SFT (MDLM)
& $73.62$
& $17.40$
& $8.98$
& $0.00$
& $18.90$
& $28.79$ \\

\midrule

Dream-Instruct
& $64.44$
& $28.20$
& $22.27$
& $6.93$
& $35.98$
& $44.36$ \\
+ SFT (MDLM)
& $70.43$
& $32.80$
& $20.31$
& $19.68$
& $37.20$
& $45.53$ \\

\midrule

Dream-Base
& $49.05$
& $20.40$
& $10.55$
& $1.07$
& $15.85$
& $22.96$ \\
+ SFT (MDLM)
& $63.00$
& $23.40$
& $8.59$
& $1.03$
& $29.88$
& $23.35$ \\

\bottomrule
\end{tabular}
\caption{\textbf{MDLM SFT evaluation results}. 
Instruct models show consistent gains, Base models gain on in-distribution math but may regress on out-of-distribution planning and coding.}
\label{tab:lora-finetuning-results}
\end{table*}

\subsection{Training Small DLMs from Scratch}
\label{sec:recipes-training-small-dlms}

In addition to finetuning open-weight large DLMs, \dllm includes \textit{recipes} and released \textit{checkpoints} for training small DLMs from scratch (starting from backbones that are not DLMs).
We cover two applications: (1) converting discriminative BERT models~\citep{devlin2019bert} into DLMs and (2) converting autoregressive LMs into DLMs~\citep{gong2024scaling}. 

\subsubsection{BERT-Chat: Converting BERTs to DLMs}
\label{sec:recipes-bert-chat}

Despite their traditional use in discriminative tasks, BERT-style models~\citep{devlin2019bert} offer bidirectional representations well-suited for diffusive generation~\citep{sahoo2024simple}. 
We show that an off-the-shelf BERT-style model can be turned into a diffusion chatbot, without architectural changes, by finetuning only on instruction-following data.
We build on top of the ModernBERT series~\citep{warner2024modernbert} as the backbone, as they are among the strongest-performing BERT variants, and release two \huggingface checkpoints, \href{https://huggingface.co/dllm-hub/ModernBERT-base-chat-v0.1}{ModernBERT-base-chat-v0.1} and \href{https://huggingface.co/dllm-hub/ModernBERT-large-chat-v0.1}{ModernBERT-large-chat-v0.1}.


\paragraph{Training details.}
We finetune ModernBERT-base and ModernBERT-large via MDLM SFT (no continual pretraining) on a mixture of instruction-tuning datasets: Tulu 3 SFT~\citep{lambert2025tulu3} and SmolTalk~\citep{allal2025smollm2}. 
The loss is computed only on response tokens.
We use maximum sequence length $1024$, $10$ epochs, learning rate $10^{-4}$, global batch size $384$, bf16 precision, and a cosine learning-rate schedule with $10\%$ warmup.
Training runs on $8 \times$A100 GPUs with DeepSpeed ZeRO-2~\citep{rajbhandari2020zero}.
See Figure~\ref{fig:recipes-bert-chat} for training curves.
We release the scripts to reproduce the models at \github \href{https://github.com/ZHZisZZ/dllm/tree/main/examples/bert}{dllm/examples/bert}.

\paragraph{Evaluation results.}
We evaluate BERT-Chats using \dllm's unified evaluation pipeline (Table~\ref{tab:bert-chat-eval}). A gap remains compared to decoder-only ARLMs of a similar size (e.g., Qwen1.5-0.5B and Qwen1.5-0.5B-Chat~\citep{bai2023qwen} on MMLU~\citep{hendrycks2021mmlu} and HellaSwag~\citep{zellers2019hellaswag}), yet the results are still noteworthy: ModernBERT-large-chat surpasses both GPT-2~\citep{radford2019language} variants on most benchmarks and outperforms Qwen1.5-0.5B-Chat on BBH~\citep{suzgun2023bbh} and MATH~\citep{hendrycks2021math}, despite being an encoder-only model with no architectural modification for generation. This suggests that BERT-style backbones are a viable, if under-explored, starting point for DLMs.

\subsubsection{Tiny-A2D: Converting ARLMs to DLMs}
\label{sec:recipes-tiny-a2d}

Autoregressive language models (ARLMs) dominate open-ended text generation, but DLMs offer complementary benefits such as parallel decoding and iterative refinement. Prior work has explored AR-to-diffusion conversion to bootstrap ARLM training artifacts into DLMs (e.g., RND1~\citep{rnd1_2025} and DiffuLLaMA~\citep{gong2024scaling}). We show that an off-the-shelf ARLM can be converted into a diffusion chatbot with minimal changes: we take Qwen3-0.6B~\citep{yang2025qwen3} as the backbone and tune it under two diffusion objectives, MDLM (masked diffusion)~\citep{sahoo2024simple} and BD3LM (block diffusion)~\citep{arriolablock}, on instruction-following data. We release two \huggingface checkpoints, \href{https://huggingface.co/dllm-hub/Qwen3-0.6B-diffusion-mdlm-v0.1}{Qwen3-0.6B-diffusion-mdlm-v0.1} and \href{https://huggingface.co/dllm-hub/Qwen3-0.6B-diffusion-bd3lm-v0.1}{Qwen3-0.6B-diffusion-bd3lm-v0.1}.

\paragraph{Training details.}
We train both variants with only SFT (no continual pretraining) on the mixture of Tulu 3 SFT~\citep{lambert2025tulu3}, SmolTalk~\citep{allal2025smollm2}, and opc-sft-stage1\&2~\citep{huang2025opencoder}. Loss is computed only on response tokens and we do not apply the logits right shifting tricks as in prior work~\citep{gong2024scaling, rnd1_2025}, because in our experiments this leads to performance degradation. For the MDLM variant we use maximum sequence length $1024$; for the BD3LM variant we use length $512$ and block size $32$. Both use $10$ epochs, learning rate $10^{-4}$, global batch size $2048$, bf16 precision, and a cosine learning-rate schedule. Training is run on $64\times$A100 GPUs with DeepSpeed ZeRO-2~\citep{rajbhandari2020zero}. See Figure~\ref{fig:recipes-tiny-a2d} for training curves.
We also release the scripts to reproduce the models at \github \href{https://github.com/ZHZisZZ/dllm/tree/main/examples/a2d}{dllm/examples/a2d}.

\paragraph{Evaluation results.}
We evaluate both converted models using \dllm's unified evaluation pipeline (Table~\ref{tab:qwen-a2d-eval}). The BD3LM variant shows particular strength on code generation, with HumanEval~\citep{chen2021humaneval} and MBPP~\citep{austin2021mbpp} scores that surpass the original Qwen3-0.6B-Base~\citep{yang2025qwen3} despite being trained with SFT alone. Overall, both DLM variants still trail their AR counterparts on most knowledge and reasoning benchmarks (e.g., MMLU~\citep{hendrycks2021mmlu}, BBH~\citep{suzgun2023bbh}), reflecting the expected gap at this scale. Nonetheless, the fact that a competitive DLM can be obtained from an off-the-shelf ARLM with only SFT and no continual pretraining demonstrates that AR-to-diffusion conversion is a practical and compute-efficient path to building DLMs.

\begin{takeaway}
\textbf{Existing pretrained models, both BERT-style encoders and autoregressive LMs, can be converted into functional DLMs with only minimal compute} (e.g., via SFT) and no architectural modification or continual pretraining, making DLM development accessible with minimal compute.
Recent work~\citep{fu2025efficientdlm} also validates that such lightweight conversion can match or surpass the DLMs trained from scratch.
\end{takeaway}

\begin{table*}[t]
    \centering
    \small
    \setlength{\tabcolsep}{2pt}
    \begin{tabular}{lccccccc}
    \toprule
    Model & GSM8K & BBH & MATH & MMLU & HellaSwag & LAMBADA & WinoGrande \\
    \midrule
    ModernBERT-base-chat-v0.1  & $3.6$  & $21.1$ & $3.1$ & $26.2$ & $34.5$ & $49.3$ & $48.8$ \\
    ModernBERT-large-chat-v0.1 & $9.3$  & $25.6$ & $3.6$ & $29.6$ & $40.9$ & $46.3$ & $49.0$ \\
    Qwen1.5-0.5B               & $22.0$ & $18.3$ & $3.1$ & $39.2$ & $48.2$ & $48.6$ & $55.0$ \\
    Qwen1.5-0.5B-Chat          & $11.3$ & $18.2$ & $2.1$ & $35.0$ & $36.9$ & $41.2$ & $52.0$ \\
    GPT-2                      & $0.7$  & $6.9$  & $1.8$ & $22.9$ & $31.1$ & $46.0$ & $51.6$ \\
    GPT-2-medium               & $2.1$  & $17.8$ & $1.4$ & $22.9$ & $39.4$ & $55.5$ & $53.1$ \\
    \bottomrule
    \end{tabular}
    \vspace{-0.5em}
    \caption{\textbf{ModernBERT-Chat evaluation results.} ModernBERT-Chat~\citep{warner2024modernbert} and GPT-2~\citep{radford2019language} models are evaluated with \dllm's pipeline; Qwen1.5~\citep{bai2023qwen} numbers are reported by original sources. See Figure~\ref{fig:recipes-bert-chat} for training curves.}
    \label{tab:bert-chat-eval}
\end{table*}

\begin{table*}[t]
    \centering
    \small
    \setlength{\tabcolsep}{2pt}
    \begin{tabular}{lcccccccc}
    \toprule
    Model & GSM8K & BBH & MATH & MMLU & MMLU-Pro & HellaSwag & HumanEval & MBPP \\
    \midrule
    Qwen3-0.6B-mdlm-v0.1   & $29.3$ & $26.7$ & $8.7$  & $40.0$ & $17.3$ & $42.1$ & $30.5$ & $29.2$ \\
    Qwen3-0.6B-bd3lm-v0.1  & $46.3$ & $26.6$ & $12.9$ & $39.1$ & $13.8$ & $39.3$ & $46.3$ & $38.2$ \\
    Qwen2.5-0.5B           & $41.6$ & $20.3$ & $19.5$ & $47.5$ & $15.7$ & $52.1$ & $30.5$ & $39.3$ \\
    Qwen3-0.6B-Base        & $59.6$ & $41.5$ & $32.4$ & $52.8$ & $24.7$ & $47.4$ & $32.3$ & $36.6$ \\
    \bottomrule
    \end{tabular}
    \vspace{-0.5em}
    \caption{\textbf{Qwen-A2D evaluation results.} MDLM and BD3LM models are evaluated with \dllm's pipeline; Autoregressive Qwen2.5/Qwen3 baselines are reported by original sources~\citep{qwen2025qwen25technicalreport,yang2025qwen3}. See Figure~\ref{fig:recipes-tiny-a2d} for training curves.}
    \label{tab:qwen-a2d-eval}
\end{table*}

\section{Related Work}
\label{sec:related-work}

\paragraph{Discrete diffusion for text.}
Diffusion models, originally developed for continuous domains~\citep{sohl2015deep,ho2020denoising,song2020score}, are extended to discrete text via absorbing-state (D3PM~\citep{austin2021structured}) and uniform-state (multinomial diffusion~\citep{hoogeboom2021argmax}) formulations. Continuous-time extensions~\citep{campbell2022continuous}, score-based~\citep{sun2022score,meng2022concrete}, and ratio-based~\citep{lou2023discrete} objectives further unify the theory. Masked diffusion language models (MDLMs) simplify the forward process to independent token masking, with recent work clarifying equivalences and simplifying training~\citep{sahoo2024simple,shi2024simplified,ou2024your,zheng2024masked}. Alternative directions include continuous diffusion in embedding space~\citep{li2022diffusion,gong2022diffuseq,dieleman2022continuous,lin2023text}, flow matching and edit-based methods~\citep{gat2024discrete,havasi2025edit,nguyen2025oneflow}, block diffusion~\citep{arriolablock}, which interpolates between AR and diffusion decoding for KV-cache reuse, and hybrid AR--diffusion architectures that use diffusion for speculative drafting~\citep{christopher2025specdiff}, planned outline-then-diffuse generation~\citep{israel2025planned}, or unified draft-and-verify passes~\citep{liu2025tidar}.

\paragraph{Open-weight DLMs.}
Scaling DLMs has progressed rapidly. \citeauthor{nie2024scaling} first scaled masked diffusion to 1.1B parameters. Converting pretrained autoregressive models into DLMs has proven effective: DiffuGPT/DiffuLLaMA adapt GPT-2 and LLaMA (127M--7B)~\citep{gong2024scaling}; RND1~\citep{rnd1_2025} extends this to 30B; and LLaDA2.0~\citep{bie2025llada2} scales to 100B with a 3-phase block-level scheme. At the 7--8B scale, Dream~\citep{ye2025dream} adapts Qwen-2.5 with context-adaptive noise rescheduling, and LLaDA~\citep{nielarge} trains an 8B MDLM from scratch, achieving performance competitive with LLaMA3-8B~\citep{grattafiori2024llama3}. Commercial systems such as Mercury~\citep{khanna2025mercury} further demonstrate DLM viability in production.

\paragraph{Open tools for DLMs.}
Prior open-weight DLMs~\citep{nielarge,ye2025dream,gong2024scaling,rnd1_2025,bie2025llada2} often lack unified development pipelines, making reproduction and comparison difficult. Open efficient inference tools for DLMs such as Fast-dLLM~\citep{wu2025fastdllm} and Fast-dLLM v2~\citep{wu2025fastdllmv2} accelerate decoding but are developed independently of training and evaluation. Evaluation pipelines also vary across papers (e.g., task sets, inference hyperparameters), and interfaces remain inconsistent. A framework unifying training, inference, and evaluation has been lacking. \dllm fills this gap with modular trainers, a plug-and-play sampler abstraction, and a reproducible evaluation pipeline aligned with official benchmarks (Section~\ref{sec:dllm-overview}, Section~\ref{sec:recipes}).

\section{Conclusion}
\label{sec:conclusion}


We present \dllm, an open-source framework that unifies the training, inference, and evaluation of DLMs in a modular, extensible pipeline. By standardizing the common components shared across recent DLMs, \dllm lowers the barrier to reproducing, finetuning, and fairly comparing existing models while making it straightforward to integrate new designs. Alongside the framework, we provide minimal recipes and checkpoints showing that existing pretrained models, both BERT-style encoders and autoregressive LMs, can be converted into competitive DLMs with lightweight finetuning alone, making DLM development increasingly accessible with minimal compute. We hope \dllm accelerates research and lowers the entry barrier for the broader community.

\paragraph{Future work.}
We plan to continue expanding \dllm by incorporating new methods as the field evolves, e.g., integrating RL algorithms once widely adopted approaches for DLMs emerge, and supporting additional open-weight models as they are released.

\section*{Acknowledgements}
Zhanhui Zhou gratefully acknowledges support from the Berkeley Fellowship.

\bibliography{colm2025_conference}
\bibliographystyle{colm2025_conference}

\appendix
\clearpage
\section{Training Curves}

\begin{figure}[htbp]
    \centering
    \includegraphics[width=0.95\linewidth]{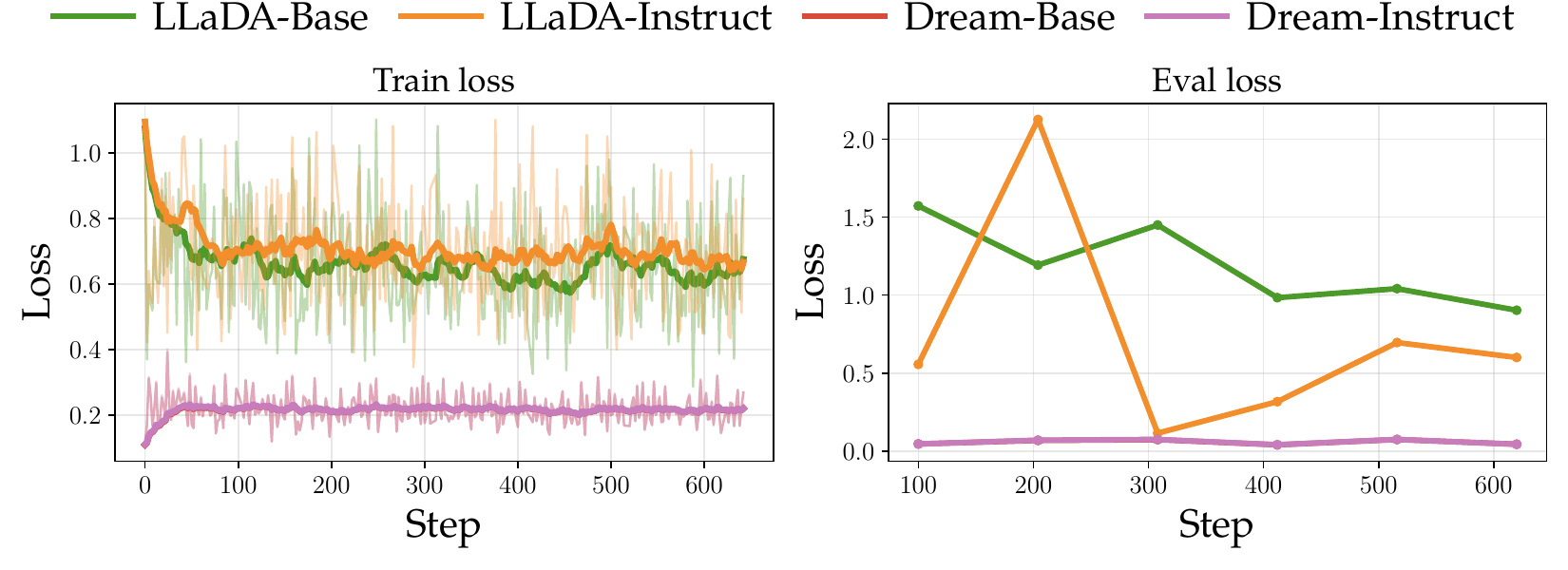}
    \vspace{-1em}
    \caption{Training loss for finetuning open-weight DLMs to reason (Section~\ref{sec:recipes-finetuning-open-weight-large-dlms}).}
    \label{fig:recipes-finetuning-open-weight-large-dlms}
\end{figure}

\vspace{2em}

\begin{figure}[htbp]
    \centering
    \includegraphics[width=0.95\linewidth]{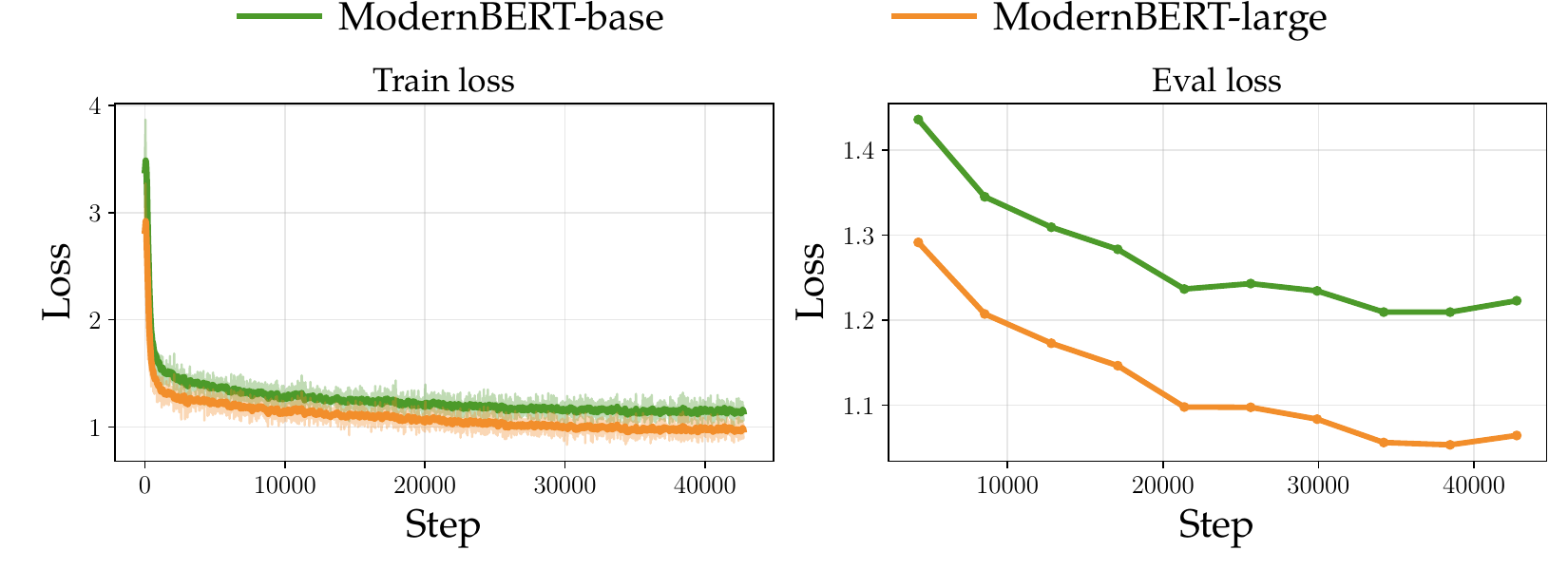}
    \vspace{-1em}
    \caption{Training loss for finetuning BERT to chat (Section~\ref{sec:recipes-bert-chat}).}
    \label{fig:recipes-bert-chat}
\end{figure}

\vspace{2em}

\begin{figure}[htbp]
    \centering
    \includegraphics[width=0.95\linewidth]{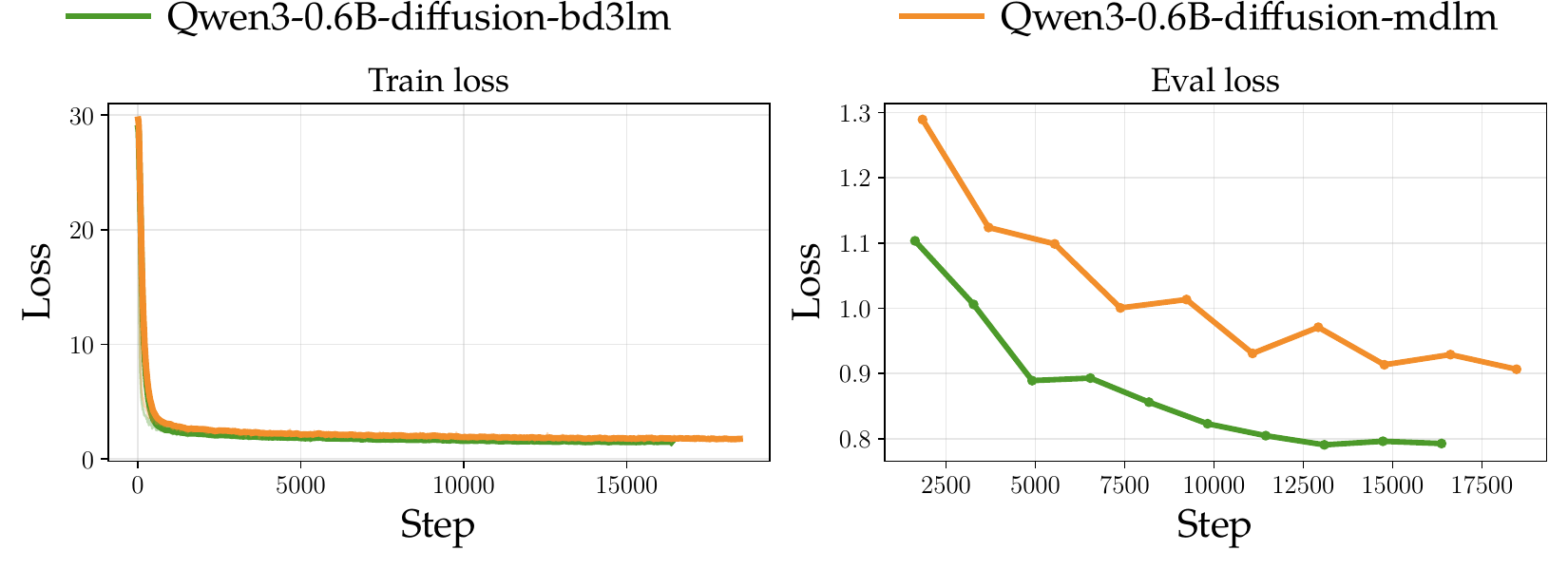}
    \vspace{-1em}
    \caption{Training loss for finetuning autoregressive LMs to be DLMs (Section~\ref{sec:recipes-tiny-a2d}).}
    \label{fig:recipes-tiny-a2d}
\end{figure}
\clearpage
\section{Evaluation Reproduction}
\label{app:sec:evaluation-reproduction}

In this section, we report evaluation results comparing the official implementation (as reported in the original paper) with our unified \dllm reimplementation under the same configurations. Overall, our framework reproduces the official results closely across benchmarks, indicating that our evaluation pipeline and implementation are consistent with the official setup (with only minor necessary adjustments).

Tables~\ref{tab:llada-eval} and~\ref{tab:dream-eval} report our reproduced evaluation results for LLaDA~\citep{nielarge} and Dream~\citep{ye2025dream} with \dllm. Tables~\ref{tab:llada-fastdllm-eval} and~\ref{tab:dream-fastdllm-eval} report Fast-dLLM results, showing that our \dllm reimplementation achieves similar accuracy to the official numbers while substantially improving generation throughput.

  
  

\begin{table*}[h]
  \centering
  \small
  \setlength{\tabcolsep}{0pt} 
  \caption{\textbf{LLaDA evaluation results.} ``Official'' denotes results from the original paper; ``\dllm'' denotes results from our \dllm reimplementation. ``Hella.'' stands for HellaSwag, ``HEval'' stands for HumanEval and ``WinoG.'' stands for WinoGrande.}
  \label{tab:llada-eval}

  \subcaption{LLaDA-Base}
  \begin{tabular*}{\textwidth}{@{\extracolsep{\fill}}lccccccccccc}
  \toprule 
  & MMLU & BBH & ARC-C & Hella. & WinoG. & PIQA 
  & GSM8K & MATH & GPQA & HEval & MBPP \\
  \midrule
  Official
  & 65.9 & 49.7 & 45.9 & 70.5 & 74.8 & 73.6 
  & 70.3 & 31.4 & 25.2 & 35.4 & 40.0 \\
  
  \dllm
  & 65.9 & 47.2 & 44.1 & 69.2 & 70.4 & 70.7 
  & 70.7 & 32.4 & 31.9 & 32.9 & 38.8 \\
  \bottomrule
  \end{tabular*}

  \vspace{1em} 

  \subcaption{LLaDA-Instruct}
  \begin{tabular*}{\textwidth}{@{\extracolsep{\fill}}lccccccccc}
  \toprule 
  & MMLU & MMLU-Pro & ARC-C & Hella. & GSM8K & Math & GPQA & HEval & MBPP \\
  \midrule
  Official
  & 65.5 & 37.0 & 88.5 & 74.6 & 69.4 & 31.9 & 33.3 & 49.4 & 41.0 \\
  
  \dllm
  & 69.8 & 36.2 & 86.4 & 76.7 & 74.7 & 31.9 & 30.6 & 47.0 & 40.0 \\
  \bottomrule
  \end{tabular*}
\end{table*}

  
  
  

\begin{table*}[h]
  \centering
  \small
  \setlength{\tabcolsep}{0pt} 
  \caption{\textbf{Dream evaluation results.} ``Official'' denotes results from the original paper; ``\dllm'' denotes results from our \dllm reimplementation. ``Hella.'' stands for HellaSwag, ``HEval'' stands for HumanEval and ``WinoG.'' stands for WinoGrande.}
  \label{tab:dream-eval}

  \subcaption{Dream-Base}
  \begin{tabular*}{\textwidth}{@{\extracolsep{\fill}}lccccccccccc}
  \toprule
  & MMLU & BBH & ARC-C & Hella. & WinoG. & PIQA 
  & GSM8K & MATH & GPQA & HEval & MBPP \\
  \midrule
  Official
  & 69.5 & 57.9 & 59.9 & 73.3 & 74.8 & 75.8 
  & 77.2 & 39.6 & 36.6 & 57.9 & 56.2 \\
  
  \dllm
  & 70.0 & 63.7 & 59.0 & 73.5 & 72.5 & 76.4 
  & 77.0 & 42.4 & 34.6 & 56.7 & 56.0 \\
  \bottomrule
  \end{tabular*}
  
  \vspace{1em} 
  
  \subcaption{Dream-Instruct}
  \begin{tabular*}{\textwidth}{@{\extracolsep{\fill}}lccccccccc}
  \toprule
  & MMLU & MMLU-Pro & ARC-C & Hella. & GSM8K & MATH & GPQA & HEval & MBPP \\
  \midrule
  Official
  & 67.0 & 43.3 & --- & --- & 81.0 & 39.2 & 33.0 & 55.5 & 58.8 \\
  
  \dllm
  & 69.8 & 45.5 & 61.4 & 71.8 & 82.0 & 48.6 & 31.5 & 57.9 & 58.2 \\
  \bottomrule
  \end{tabular*}
\end{table*}

\begin{table*}[h]
    \centering
    \small
    \caption{\textbf{Fast-dLLM LLaDA-Instruct evaluation results with max new tokens @ $256$ (a) and $512$ (b).} ``Official'' denotes results from the Fast-dLLM paper; ``\dllm'' denotes results from our \dllm reimplementation.}
    \label{tab:llada-fastdllm-eval}
    \subcaption{max new tokens @ $256$}
    \begin{tabular}{l c | c c c c c c c c}
    \toprule
    \textbf{Benchmark} & \textbf{Source} &
    \multicolumn{2}{c}{\textbf{Baseline}} &
    \multicolumn{2}{c}{\textbf{+Cache}} &
    \multicolumn{2}{c}{\textbf{+Parallel}} &
    \multicolumn{2}{c}{\textbf{+Both}} \\
    & & Acc & {\color{blue}Tok/s ($\times$)} 
    & Acc & {\color{blue}($\times$)} 
    & Acc & {\color{blue}($\times$)} 
    & Acc & {\color{blue}($\times$)} \\
    \midrule
    
    GSM8K & Official 
    & $79.3$ & {\color{blue}$6.7\ (1.0 \times)$}
    & $79.5$ & {\color{blue}$3.2 \times$}
    & $79.2$ & {\color{blue}$2.5 \times$}
    & $78.5$ & {\color{blue}$8.1 \times$} \\
    & \dllm
    & $78.0$ & {\color{blue}$8.1\ (1.0 \times)$}
    & $78.2$ & {\color{blue}$3.2 \times$}
    & $78.9$ & {\color{blue}$2.3 \times$}
    & $78.0$ & {\color{blue}$6.5 \times$} \\
    \midrule
    
    MATH & Official 
    & $33.5$ & {\color{blue}$9.1\ (1.0 \times)$}
    & $33.3$ & {\color{blue}$2.6 \times$}
    & $33.4$ & {\color{blue}$2.7 \times$}
    & $33.2$ & {\color{blue}$5.7 \times$} \\
    & \dllm
    & $38.3$ & {\color{blue}$9.7\ (1.0 \times)$}
    & $37.6$ & {\color{blue}$2.7 \times$}
    & $38.6$ & {\color{blue}$2.0 \times$}
    & $37.5$ & {\color{blue}$5.0 \times$} \\
    \midrule
    
    HumanEval & Official 
    & $41.5$ & {\color{blue}$30.5\ (1.0 \times)$}
    & $42.7$ & {\color{blue}$1.3 \times$}
    & $43.9$ & {\color{blue}$3.3 \times$}
    & $43.3$ & {\color{blue}$3.7 \times$} \\
    & \dllm
    & $38.4$ & {\color{blue}$18.8\ (1.0 \times)$}
    & $36.0$ & {\color{blue}$1.5 \times$}
    & $39.6$ & {\color{blue}$2.8 \times$}
    & $36.0$ & {\color{blue}$3.6 \times$} \\
    \midrule
    
    MBPP & Official 
    & $29.4$ & {\color{blue}$6.0\ (1.0 \times)$}
    & $29.6$ & {\color{blue}$2.8 \times$}
    & $28.4$ & {\color{blue}$4.1 \times$}
    & $28.2$ & {\color{blue}$7.5 \times$} \\
    & \dllm
    & $36.4$ & {\color{blue}$9.3\ (1.0 \times)$}
    & $38.0$ & {\color{blue}$2.8 \times$}
    & $29.0$ & {\color{blue}$1.9 \times$}
    & $37.8$ & {\color{blue}$4.8 \times$} \\
    
    \bottomrule
    \end{tabular}
    \vspace{1em}
    \subcaption{max new tokens @ $512$}
    \begin{tabular}{l c | c c c c c c c c}
    \toprule
    \textbf{Benchmark} & \textbf{Source} &
    \multicolumn{2}{c}{\textbf{Baseline}} &
    \multicolumn{2}{c}{\textbf{+Cache}} &
    \multicolumn{2}{c}{\textbf{+Parallel}} &
    \multicolumn{2}{c}{\textbf{+Both}} \\
    & & Acc & {\color{blue}Tok/s ($\times$)} 
    & Acc & {\color{blue}($\times$)} 
    & Acc & {\color{blue}($\times$)} 
    & Acc & {\color{blue}($\times$)} \\
    \midrule

    GSM8K & Official 
    & $77.5$ & {\color{blue}$3.2\ (1.0 \times)$}
    & $77.0$ & {\color{blue}$3.3 \times$}
    & $77.6$ & {\color{blue}$5.8 \times$}
    & $77.2$ & {\color{blue}$11.0 \times$} \\
    & \dllm
    & $81.1$ & {\color{blue}$6.7\ (1.0 \times)$}
    & $76.0$ & {\color{blue}$3.0 \times$}
    & $77.6$ & {\color{blue}$3.3 \times$}
    & $76.6$ & {\color{blue}$7.8 \times$} \\
    \midrule

    MATH & Official 
    & $37.2$ & {\color{blue}$8.0\ (1.0 \times)$}
    & $36.2$ & {\color{blue}$2.5 \times$}
    & $36.8$ & {\color{blue}$3.0 \times$}
    & $36.0$ & {\color{blue}$5.9 \times$} \\
    & \dllm
    & $42.4$ & {\color{blue}$7.4\ (1.0 \times)$}
    & $41.9$ & {\color{blue}$2.9 \times$}
    & $42.5$ & {\color{blue}$2.7 \times$}
    & $41.8$ & {\color{blue}$6.0 \times$} \\
    \midrule

    HumanEval & Official 
    & $43.9$ & {\color{blue}$18.4\ (1.0 \times)$}
    & $45.7$ & {\color{blue}$1.6 \times$}
    & $43.3$ & {\color{blue}$3.1 \times$}
    & $44.5$ & {\color{blue}$4.0 \times$} \\
    & \dllm
    & $48.2$ & {\color{blue}$13.0\ (1.0 \times)$}
    & $41.5$ & {\color{blue}$1.8 \times$}
    & $50.6$ & {\color{blue}$2.8 \times$}
    & $41.5$ & {\color{blue}$4.3 \times$} \\
    \midrule

    MBPP & Official 
    & $14.8$ & {\color{blue}$4.3\ (1.0 \times)$}
    & $13.4$ & {\color{blue}$2.3 \times$}
    & $15.0$ & {\color{blue}$5.1 \times$}
    & $13.8$ & {\color{blue}$9.2 \times$} \\
    & \dllm
    & $32.2$ & {\color{blue}$7.7\ (1.0 \times)$}
    & $22.0$ & {\color{blue}$2.7 \times$}
    & $7.6$  & {\color{blue}$2.7 \times$}
    & $21.4$ & {\color{blue}$5.7 \times$} \\

    \bottomrule
    \end{tabular}
\end{table*}

\begin{table*}[t]
    \centering
    \small
    \caption{\textbf{Fast-dLLM Dream-Base evaluation results with max new tokens @ $256$ (a) and $512$ (b).} ``Official'' denotes results from the Fast-dLLM paper; ``\dllm'' denotes results from our \dllm reimplementation.}
    \label{tab:dream-fastdllm-eval}
    \subcaption{max new tokens @ $256$}
    \begin{tabular}{l c | c c c c c c c c}
    \toprule
    \textbf{Benchmark} & \textbf{Source} &
    \multicolumn{2}{c}{\textbf{Baseline}} &
    \multicolumn{2}{c}{\textbf{+Cache}} &
    \multicolumn{2}{c}{\textbf{+Parallel}} &
    \multicolumn{2}{c}{\textbf{+Both}} \\
    & & Acc & {\color{blue}Tok/s ($\times$)} 
    & Acc & {\color{blue}($\times$)} 
    & Acc & {\color{blue}($\times$)} 
    & Acc & {\color{blue}($\times$)} \\
    \midrule
    
    GSM8K & Official
    & $75.0$ & {\color{blue}$9.1\ (1.0 \times)$}
    & $74.3$ & {\color{blue}$3.6 \times$}
    & $74.2$ & {\color{blue}$1.6 \times$}
    & $74.8$ & {\color{blue}$5.3 \times$} \\
    & \dllm
    & $75.4$ & {\color{blue}$9.0\ (1.0 \times)$}
    & $75.0$ & {\color{blue}$3.7 \times$}
    & $72.6$ & {\color{blue}$1.4 \times$}
    & $74.2$ & {\color{blue}$4.7 \times$} \\
    \midrule
    
    MATH & Official
    & $38.4$ & {\color{blue}$11.4\ (1.0 \times)$}
    & $36.8$ & {\color{blue}$3.0 \times$}
    & $37.9$ & {\color{blue}$2.4 \times$}
    & $37.6$ & {\color{blue}$5.9 \times$} \\
    & \dllm
    & $31.5$ & {\color{blue}$25.1\ (1.0 \times)$}
    & $33.3$ & {\color{blue}$1.5 \times$}
    & $23.5$ & {\color{blue}$2.1 \times$}
    & $31.1$ & {\color{blue}$3.1 \times$} \\
    \midrule
    
    HumanEval & Official
    & $49.4$ & {\color{blue}$23.3\ (1.0 \times)$}
    & $53.7$ & {\color{blue}$1.5 \times$}
    & $49.4$ & {\color{blue}$2.0 \times$}
    & $54.3$ & {\color{blue}$2.8 \times$} \\
    & \dllm
    & $57.9$ & {\color{blue}$14.0\ (1.0 \times)$}
    & $53.7$ & {\color{blue}$2.4 \times$}
    & $51.2$ & {\color{blue}$1.5 \times$}
    & $53.1$ & {\color{blue}$3.1 \times$} \\
    \midrule
    
    MBPP & Official
    & $56.6$ & {\color{blue}$11.2\ (1.0 \times)$}
    & $53.2$ & {\color{blue}$3.1 \times$}
    & $53.8$ & {\color{blue}$2.8 \times$}
    & $56.4$ & {\color{blue}$6.8 \times$} \\
    & \dllm
    & $55.6$ & {\color{blue}$9.9\ (1.0 \times)$}
    & $53.8$ & {\color{blue}$3.3 \times$}
    & $53.6$ & {\color{blue}$2.5 \times$}
    & $56.0$ & {\color{blue}$6.3 \times$} \\
    
    \bottomrule
    \end{tabular}
    \vspace{1em}
    \subcaption{max new tokens @ $512$}
    \begin{tabular}{l c | c c c c c c c c}
    \toprule
    \textbf{Benchmark} & \textbf{Source} &
    \multicolumn{2}{c}{\textbf{Baseline}} &
    \multicolumn{2}{c}{\textbf{+Cache}} &
    \multicolumn{2}{c}{\textbf{+Parallel}} &
    \multicolumn{2}{c}{\textbf{+Both}} \\
    & & Acc & {\color{blue}Tok/s ($\times$)} 
    & Acc & {\color{blue}($\times$)} 
    & Acc & {\color{blue}($\times$)} 
    & Acc & {\color{blue}($\times$)} \\
    \midrule
    
    GSM8K & Official
    & $76.0$ & {\color{blue}$7.7\ (1.0 \times)$}
    & $74.3$ & {\color{blue}$3.3 \times$}
    & $73.4$ & {\color{blue}$1.9 \times$}
    & $74.0$ & {\color{blue}$5.6 \times$} \\
    & \dllm
    & $75.7$ & {\color{blue}$7.6\ (1.0 \times)$}
    & $73.8$ & {\color{blue}$3.4 \times$}
    & $72.7$ & {\color{blue}$1.6 \times$}
    & $74.5$ & {\color{blue}$4.4 \times$} \\
    \midrule
    
    MATH & Official
    & $39.8$ & {\color{blue}$9.6\ (1.0 \times)$}
    & $38.0$ & {\color{blue}$2.8 \times$}
    & $39.5$ & {\color{blue}$3.2 \times$}
    & $39.3$ & {\color{blue}$6.5 \times$} \\
    & \dllm
    & $39.2$ & {\color{blue}$15.8\ (1.0 \times)$}
    & $39.2$ & {\color{blue}$1.8 \times$}
    & $32.0$ & {\color{blue}$1.6 \times$}
    & $38.9$ & {\color{blue}$2.9 \times$} \\
    \midrule
    
    HumanEval & Official
    & $54.3$ & {\color{blue}$16.3\ (1.0 \times)$}
    & $54.9$ & {\color{blue}$1.7 \times$}
    & $51.8$ & {\color{blue}$1.8 \times$}
    & $54.3$ & {\color{blue}$3.2 \times$} \\
    & \dllm
    & $54.9$ & {\color{blue}$10.4\ (1.0 \times)$}
    & $54.9$ & {\color{blue}$2.5 \times$}
    & $50.6$ & {\color{blue}$1.6 \times$}
    & $54.3$ & {\color{blue}$3.6 \times$} \\
    \midrule
    
    MBPP & Official
    & $55.6$ & {\color{blue}$9.4\ (1.0 \times)$}
    & $53.8$ & {\color{blue}$2.8 \times$}
    & $55.4$ & {\color{blue}$4.0 \times$}
    & $55.2$ & {\color{blue}$7.8 \times$} \\
    & \dllm
    & $56.0$ & {\color{blue}$4.6\ (1.0 \times)$}
    & $52.6$ & {\color{blue}$5.4 \times$}
    & $52.8$ & {\color{blue}$6.4 \times$}
    & $54.4$ & {\color{blue}$13.3 \times$} \\
    
    \bottomrule
    \end{tabular}
\end{table*}

\end{document}